\documentclass[letterpaper, 10 pt, journal, twoside]{IEEEtran}
\ifCLASSINFOpdf
\else
\fi
\usepackage{graphics} 
\usepackage{epsfig} 
\usepackage{amsmath} 
\usepackage{amssymb}  
\usepackage{ifpdf}
\usepackage{algorithm}
\usepackage{algpseudocode}
\usepackage{multirow}
\usepackage[hidelinks]{hyperref}
\usepackage{fancyhdr}
\usepackage[caption=false, font=footnotesize]{subfig}
\usepackage{xcolor}
\usepackage{booktabs}       
\usepackage{threeparttable} 

\begin{document}
%
\title{SIS: Seam-Informed Strategy for T-shirt Unfolding}
%
%
%

\author{Xuzhao Huang$^{1}$~\IEEEmembership{Graduate Student Member,~IEEE}, Akira Seino$^{1}$~\IEEEmembership{Member,~IEEE}, \\Fuyuki Tokuda$^{1}$~\IEEEmembership{Member,~IEEE}, Akinari Kobayashi$^{1}$~\IEEEmembership{Member,~IEEE}, \\Dayuan Chen$^{2}$~\IEEEmembership{Graduate Student Member,~IEEE}, Yasuhisa Hirata$^{2}$~\IEEEmembership{Member,~IEEE}, \\Norman C. Tien$^{3}$~\IEEEmembership{Senior Member,~IEEE}, and Kazuhiro Kosuge$^{1}$~\IEEEmembership{Life Fellow,~IEEE}
\thanks{Manuscript received: March 7, 2025; Accepted May 12, 2025.}
\thanks{This paper was recommended for publication by Associate Editor S. Foix and Editor J. B. Sol upon evaluation of the Associate Editor and Reviewers' comments.
This work was supported in part by the JC STEM Lab of Robotics for Soft Materials funded by The Hong Kong Jockey Club Charities Trust and in part by the Innovation and Technology Commission of the HKSAR Government under the InnoHK initiative. } 
\thanks{$^{1}$Xuzhao Huang, Akira Seino, Fuyuki Tokuda, Akinari Kobayashi, and Kazuhiro Kosuge are with the JC STEM Lab of Robotics for Soft Materials, Department of Electrical and Electronic Engineering, Faculty of Engineering, The University of Hong Kong, Hong Kong SAR, 000000, China, and also with the Centre for Transformative Garment Production, Hong Kong SAR, 000000, China (e-mail: x.z.huang@connect.hku.hk).}%
\thanks{$^{2}$Dayuan Chen and Yasuhisa Hirata are with the Centre for Transformative Garment Production, Hong Kong SAR, 000000, China, and also with the Hirata Laboratory, Department of Robotics, Graduate School of Engineering, Tohoku University, Sendai, 980-8579, Japan. }%
\thanks{$^{3}$Norman C. Tien is with the Centre for Transformative Garment Production, Hong Kong SAR, 000000, China, and also with the Department of Electrical and Electronic Engineering, Faculty of Engineering, The University of Hong Kong, Hong Kong SAR, 000000, China. }%
\thanks{Digital Object Identifier (DOI): 10.1109/LRA.2025.3574966.}
}

%
%

\markboth{IEEE Robotics and Automation Letters. Preprint Version. Accepted May, 2025}
{Huang \MakeLowercase{\textit{et al.}}: SIS: Seam-Informed Strategy for T-shirt Unfolding}

%



\maketitle

\begin{abstract}
Seams are information-rich components of garments. 
The presence of different types of seams and their combinations helps to select grasping points for garment handling. 
In this paper, we propose a new Seam-Informed Strategy (SIS) for finding actions for handling a garment, such as grasping and unfolding a T-shirt.
Candidates for a pair of grasping points for a dual-arm manipulator system are extracted using the proposed Seam Feature Extraction Method (SFEM).
A pair of grasping points for the robot system is selected by the proposed Decision Matrix Iteration Method (DMIM). 
The decision matrix is first computed by multiple human demonstrations and updated by the robot execution results to improve the grasping and unfolding performance of the robot. 
Note that the proposed scheme is trained on real data without relying on simulation. 
Experimental results demonstrate the effectiveness and generalization ability of the proposed strategy. 
The project home page is available at {https://github.com/lancexz/sis}
\end{abstract}


%
\IEEEpeerreviewmaketitle


\section{Introduction}\label{sec:intro}
Garment unfolding remains an open challenge in robotics. 
Existing research utilizes folds~\cite{JunctionsHorizon, foldingCloth, folds}, edges~\cite{realfakeedge, VisibleConnectivity}, outline points, and structural regions~\cite{graspcollar, UniGarmentManip} as primary references for selecting grasping points, or uses a value map calculated using models trained in simulators~\cite{flingbot, clothfunnels, updownoriented, unifolding, fabricFolding}. 
However, none of these methods have utilized seam information. 
Seams are usually located in the contour position of the garment when fully unfolded.
Consequently, the desirable grasping points for unfolding the garment tend to fall near the seams. 

We believe that the introduction of seam information could improve the efficiency of the garment unfolding process for the following reasons: 
\begin{itemize}
    \item Seams can be used as a universal garment feature due to their prevalence in different types of garments. 
    \item Seams are more visible than other features when the garment is randomly placed, facilitating the perception process without additional reconfiguration of the garment.
    \item The introduction of seam information makes it possible to select the grasping points without explicitly using the garment structure, resulting in efficient garment handling.  
\end{itemize}

\begin{figure}[tbp]
    \includegraphics[width=\linewidth]{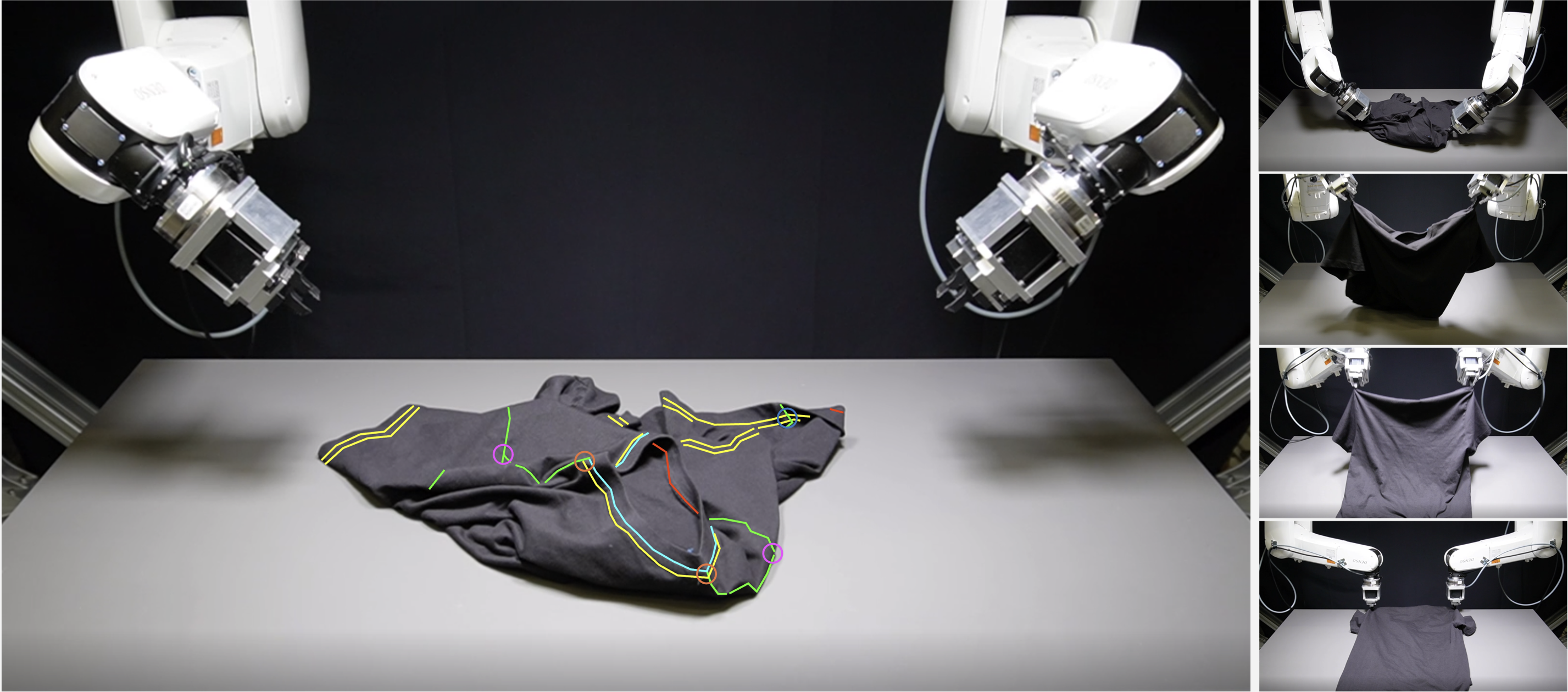}
    \caption{
    T-shirt unfolding strategy based on seam features constructed from seam line segments and their crossings. 
    The candidates for a pair of grasping points for a robot motion primitive are limited to the seam features marked in the figure. 
    }
    \label{fig:firstchart}
\end{figure}

This paper presents Seam-Informed Strategy (SIS), a novel scheme for selecting actions that can be used for automatic garment handling, such as dual-arm robotic T-shirt unfolding. 
We use the seam segments and their crossings as reference information to limit the search space for selecting grasping points, as shown in Fig.~\ref{fig:firstchart}. 

To facilitate the SIS, we propose a Seam Feature Extraction Method (SFEM) and a Decision Matrix Iteration Method (DMIM).
The SFEM is used to extract seam features as candidates for a pair of dual-arm grasping points. 
The DMIM is used as a comprehensive grasping points selection method from the extracted candidates for dual-arm unfolding motion planning.
This motion involves a sequence of robot motion primitives including grasping, stretching, and flinging to flatten and align the T-shirt.

We train the networks in SFEM with real data annotated by humans. 
The decision matrix in DMIM is initially computed with human demonstrations and updated with real robot execution results.
Using the decision matrix, the robot can also align the orientation of the unfolded T-shirt.
We evaluate the efficiency of our scheme on the real robot system through extensive experiments. 

In summary, our contributions include:
\begin{itemize}
    \item We propose SIS, a novel strategy for selecting grasping points for T-shirt unfolding using the seam information.
    SFEM and DMIM are proposed for this strategy.
    \item In the proposed SFEM, we formulate the extraction of seam lines as an oriented line object detection problem. 
    This formulation allows the use of any object detection network to efficiently handle curved/straight seam lines.
    \item We solve the grasping points selection as a scoring problem for combinations of seam segment types using a proposed DMIM, a low-cost solution for unfolding a garment. 
    \item Experimental results demonstrate that the performance of unfolding the T-shirt is promising in terms of obtaining high evaluation metrics with few episode steps.
\end{itemize}


\section{Related Work}\label{sec:relatedwork}
Unfolding and flattening is the first process that enables various garment manipulation tasks. 
However, the selection of grasping points remains a challenging aspect of garment unfolding.
Much work has been done so far.  

\subsection{Grasping Points Selection Strategies}
There are three main types of strategies for selecting grasping points in garment unfolding or other garment handling tasks: heuristic-based strategy, matching-based strategy, and value-map-based strategy. 

\subsubsection{Heuristic-based Strategy}
A common way is to limit the possible garment configurations by first randomly picking up the garment and then grasping the lowest point~\cite{realfakeedge, 3stepunfolding, activeRandomForests}. 
A heuristic method proposed by~\cite{3stepunfolding} is to unfold a garment by repeatedly grasping the lowest point of the picked up garment. 

\subsubsection{Matching-based Strategy} This strategy first maps the observed garment configuration to a canonical configuration and then obtains the grasping points based on the canonical configuration~\cite{GarmentNets}. 
The method proposed in~\cite{RTposeestimation} estimates the garment pose and the corresponding grasping point by searching a database constructed for each garment type. 

\subsubsection{Value-map-based Strategy} Recently, 
self-supervised learning frameworks~\cite{flingbot, clothfunnels, updownoriented, unifolding, fabricFolding, speedfolding} have been used to predict action value maps or ranking scores that indicate the unfolding performance of grasping point candidates. 
\cite{clothfunnels, updownoriented, unifolding} also consider the resulting garment orientation from the prioritized action.

\subsection{Datasets for Learning-based Methods}
One of the key issues for learning-based methods is the lack of data sets when dealing with garments due to the difficulty of annotation. 
To address this problem, most researchers~\cite{flingbot, clothfunnels, updownoriented, unifolding, fabricFolding, EMDNet} train their networks using simulators.
SpeedFolding~\cite{speedfolding} uses real data annotated by humans together with self-supervised learning.
In~\cite{realfakeedge} and~\cite{graspcollar}, the use of color-marked garments is proposed to automatically generate annotated depth maps. 
The use of transparent fluorescent paints and ultraviolet (UV) light is proposed by~\cite{autobag} to generate RGB images annotated by the paints observed under UV light.

\subsection{Unfolding Actions Strategies}
In addition to the commonly used Pick\&Place, the Flinging proposed by FlingBot\cite{flingbot} is widely used for unfolding actions. 
FabricFolding\cite{fabricFolding} introduces a Pick\&Drag action to improve the unfolding of sleeves. 
Meanwhile, DextAIRity\cite{dextairity} proposes the use of airflow provided by a blower to indirectly unfold the garment, thereby reducing the workspace of robots. 
UniFolding\cite{unifolding} uses a Drag\&Mop action to reposition the garment when the grasping points are out of the robot's reach.


\section{Problem Statement}
This paper focuses on automating the process of unfolding a T-shirt from arbitrary configurations by a sequence of grasping, stretching, and flinging motions using a dual-arm manipulator system under the following assumptions:
\begin{itemize}
    \item Seams are always present on the target T-shirt. 
    \item The shirt is not in an inside-out configuration. 
\end{itemize}

The desired outcomes include reducing the number of necessary steps of robot actions (episode steps) to unfold a T-shirt and increasing the normalized coverage of the final unfolded garment. 
We also consider the final orientation of the unfolded garment. 
This section describes the problem to be solved in this paper. 

\begin{figure*}[htbp]
    \centering
    \includegraphics[width=\linewidth]{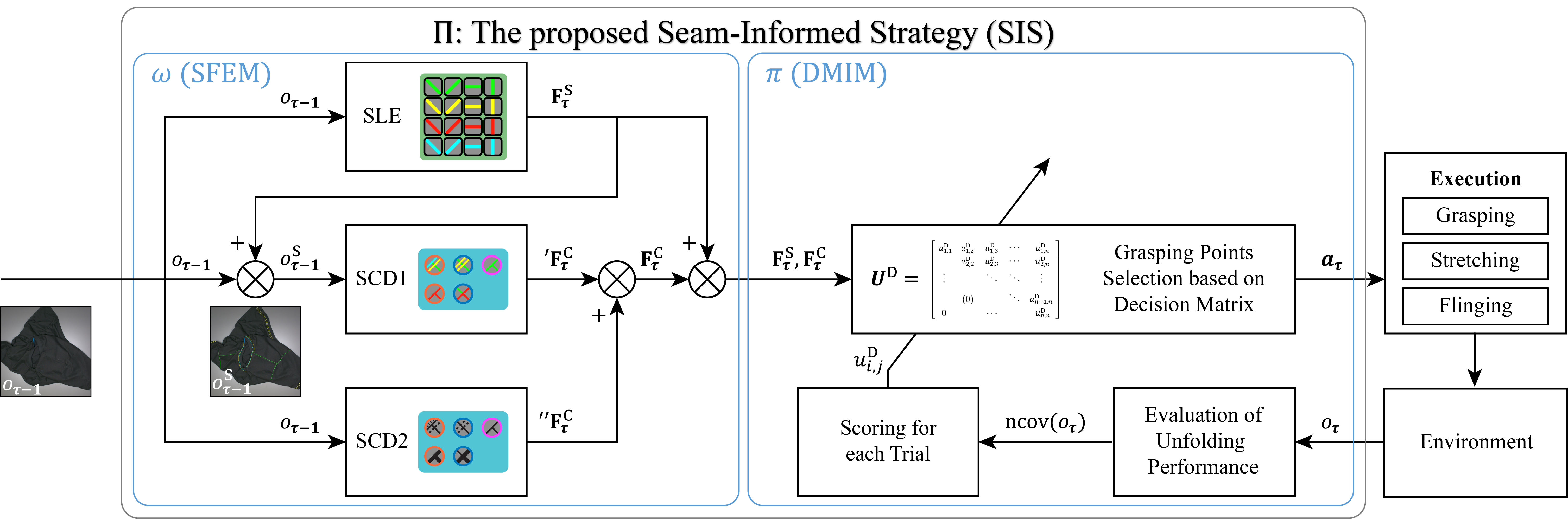}
\caption{
    The outline of the proposed SIS for unfolding a T-shirt from a randomly initialized configuration. 
    $\omega$ and $\pi$ are the methods introduced in~\ref{sec:outline} using the proposed Seam Feature Extraction Method (SFEM) and Decision Matrix Iteration Method (DMIM). 
    Given an input of an original image observation $o_{\tau-1}$, SFEM, which consists of a Seam Line-segment Extractor (SLE) and two Seam Crossing-segment Detectors (SCD1 and SCD2), extract the seam line segment set $\textbf{F}^{\mathrm{S}}_\tau$ and the seam crossing segment set $\textbf{F}^{\mathrm{C}}_\tau$ based on YOLOv3~\cite{yolo}.
    A seam-map $o_{\tau-1}^\mathrm{S}$ is generated by superimposing the extracted $\textbf{F}^{\mathrm{S}}_\tau$ onto $o_{\tau-1}$. 
    DMIM selects the grasping points according to the scoring based on a decision matrix $\textbf{\textit{U}}^{\mathrm{D}}$, and outputs an $a_\tau$, which is described by two grasping points $^{\mathrm{img}}\textbf{\textit{p}}^{\mathrm{L}}_\tau$ and $^{\mathrm{img}}\textbf{\textit{p}}^{\mathrm{R}}_\tau$. 
    After performing the action $a_\tau$ with the motion primitives, grasping, stretching, and flinging, $\textbf{\textit{U}}^{\mathrm{D}}$ is updated using the normalized coverage of the T-shirt $\mathrm{ncov}(o_\tau)$ in the observation $o_\tau$. 
    }
    \label{fig:flowchart}
\end{figure*}

\subsection{Problem Formulation}
\label{sec:problem}
Given an observation $o_{\tau-1} \in \mathbb{R}^{W \times H \times 3}$, captured by an RGB camera with resolution of $W \times H$ at episode step $\tau \in \{1,2,...\}$, our goal is to develop a policy $\Pi$ mapping $o_{\tau-1}$ to an action $a_\tau$, i.e. $a_\tau=\Pi(o_{\tau-1})$, to make the garment converge to a flattened configuration. 
The action $a_\tau$ can be described by the pair of grasping points for the left and right hands,  $^{\mathrm{img}}\textbf{\textit{p}}^{\mathrm{L}}_\tau \in \mathbb{R}^{2}$ and $ ^{\mathrm{img}}\textbf{\textit{p}}^{\mathrm{R}}_\tau \in \mathbb{R}^{2}$, w.r.t the image frame $\Sigma_{\mathrm{img}}$.

\subsection{Outline of the Proposed Scheme}
\label{sec:outline}
Existing strategies try to map the $o_{\tau-1}$ to a canonical space or a value map to select grasping points pixel-wise.
This paper proposes a new Seam-Informed Strategy (SIS) to select grasping points based on seam information. 
To implement the mapping policy $\Pi$ based on SIS, we divide the policy $\Pi$ into two methods, $\omega$ and $\pi$, as shown in Fig.~\ref{fig:flowchart}.

The method $\omega$ called the Seam Feature Extraction Method (SFEM) is proposed to extract seam line segments and their crossings from $o_{\tau-1}$.
We use the extracted set of seam line segments $\textbf{F}^{\mathrm{S}}_\tau$ and the set of seam crossing segments $\textbf{F}^{\mathrm{C}}_\tau$ as candidates for grasping points. The method $\omega$ is formulated as follows:
\begin{equation}
    \omega(o_{\tau-1}) = (\textbf{F}^{\mathrm{S}}_\tau, \textbf{F}^{\mathrm{C}}_\tau)
    \label{equ:featureExtration}
\end{equation}

The method $\pi$ called the Decision Matrix Iteration Method (DMIM) is proposed to select a pair of grasping points from the candidates obtained above for the unfolding action of the dual-arm robot. 
The decision matrix takes into account both human prior knowledge and the results of real robot executions. The method $\pi$ is formulated as follows:
\begin{equation}
    \pi(\textbf{F}^{\mathrm{S}}_\tau, \textbf{F}^{\mathrm{C}}_\tau) = (^{\mathrm{img}}\textbf{\textit{p}}^{\mathrm{L}}_\tau, ^{\mathrm{img}}\textbf{\textit{p}}^{\mathrm{R}}_\tau)
    \label{equ:graspingStrategy}
\end{equation}

From (\ref{equ:featureExtration}) and (\ref{equ:graspingStrategy}), we have: 
\begin{equation}
    \Pi({o_{\tau-1}}) = \pi(\omega(o_{\tau-1}))  = (^{\mathrm{img}}\textbf{\textit{p}}^{\mathrm{L}}_\tau, ^{\mathrm{img}}\textbf{\textit{p}}^{\mathrm{R}}_\tau)
\end{equation}


\section{Method}\label{sec:method}
As outlined in Section~\ref{sec:outline}, this paper primarily addresses two challenges. In Section~\ref{subsec:seamExtraction}, we will explain how to extract sets of grasping point candidates $\textbf{F}^{\mathrm{S}}_\tau$ and $\textbf{F}^{\mathrm{C}}_\tau$ from an RGB image input $o_{\tau-1}$ using the proposed SFEM at each episode step $\tau$. 
In Section~\ref{subsec:decisionMatrix}, we will explain how to select a pair of grasping points $^{\mathrm{img}}\textbf{\textit{p}}^{\mathrm{L}}_\tau$ and $^{\mathrm{img}}\textbf{\textit{p}}^{\mathrm{R}}_\tau$ from the candidates $\textbf{F}^{\mathrm{S}}_\tau$ and $\textbf{F}^{\mathrm{C}}_\tau$ using the proposed DMIM at each episode step $\tau$.

\subsection{Seam Feature Extraction Method (SFEM)}
\label{subsec:seamExtraction}
The proposed SFEM consists of a seam line-segment extractor (SLE) and two seam crossing-segment detectors (SCD1 and SCD2) as shown in Fig.~\ref{fig:flowchart} which are designed based on the YOLOv3~\cite{yolo}. 
Extracting seams as segments allows them to be used as grasping point candidates. 

\subsubsection{Seam Line-segment Extractor (SLE)}
The YOLOv3 was originally designed for object detection. 
For the SLE, we formulate the extraction of curved$/$straight seams as an oriented line object detection problem since the YOLOv3 cannot be used directly for seam line segment extraction. 
The proposed formulation allows any object detection network to extract seam line features. 

\begin{figure*}[htbp]
    \centering
    \includegraphics[width=\linewidth]{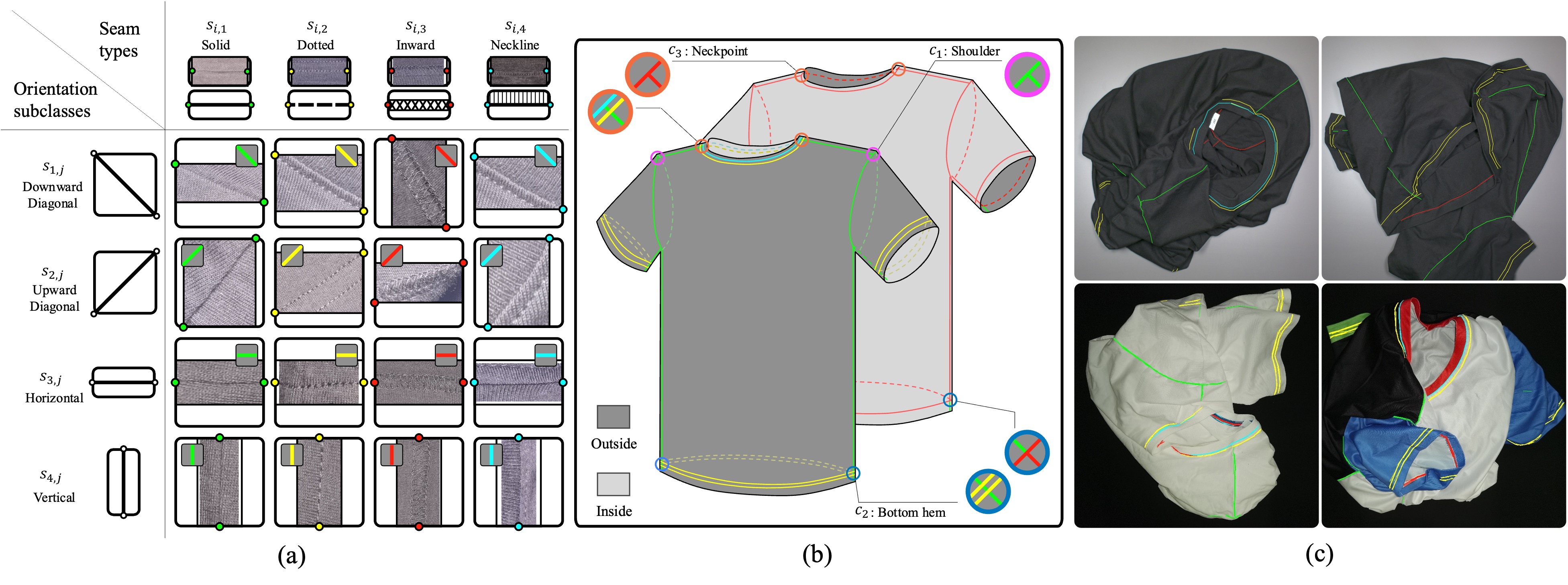}
    \caption{Categorization of the grasping point candidates in the proposed SFEM. 
    The four different colored lines represent four different types of seam segments: solid, dotted, inward, and neckline. 
    (a) The categorization method for orientated seam line objects. 
    To extract seam segments with orientation information, $s_{i,j}$ is used to denote a two-dimensional categorization for SLE in algorithm~\ref{alg:QOL}. 
    After dividing the seams into four types (solid, dotted, inward, and neckline), we further divide each type into four subclasses based on orientation (downward diagonal, upward diagonal, horizontal, and vertical). 
    (b) The distribution of each type of seam segment and the definition of their crossings in canonical space. 
    In this figure, solid lines denote visible seams, while dashed lines denote occluded seams. 
    Note that the seam structure shown here is only an example. 
    T-shirts with different seam structures are included in the training dataset. 
    (c) Visualization examples of SFEM output on real T-shirts. 
    The visualized images are named seam-maps in this paper. }
    \label{fig:subclass}
\end{figure*}

To extract seams using the object detection network, we first approximate the continuous curved seam as a set of straight line segments. 
Each straight line segment $\textbf{\textit{f}}^{\mathrm{SL}}$ is described by the seam segment category $j \in \{1,2,3,4\}$ and its endpoint positions $(x_1, y_1)$ and $(x_2, y_2)$ w.r.t the image frame $\Sigma_{\mathrm{img}}$, i.e. $\textbf{\textit{f}}^{\mathrm{SL}}=[j, x_1, y_1, x_2, y_2]$.  
In this paper, the seam segments are categorized into four categories, namely solid ($j=1$), dotted ($j=2$), inward ($j=3$), and neckline ($j=4$), as shown in Fig.~\ref{fig:subclass}~(a) and (b), based on the seam types of the T-shirt used in the experiments. 

We then convert the straight line segment $\textbf{\textit{f}}^{\mathrm{SL}}$ into a bounding box $\textbf{\textit{f}}^{\mathrm{S}}$ by using the two endpoints as the diagonal vertices of the bounding box. 
To represent the orientation of the straight line segment $\textbf{\textit{f}}^{\mathrm{SL}}$ in the bounding box $\textbf{\textit{f}}^{\mathrm{S}}$, we divide each seam segment category into four orientation subclasses $i \in \{1,2,3,4\}$, namely downward diagonal ($i=1$), upward diagonal ($i=2$), horizontal ($i=3$), and vertical ($i=4$), as shown in Fig.~\ref{fig:subclass}~(a). 
This quadrupled the number of seam segment categories.
Consequently, each oriented seam line segment $\textbf{\textit{f}}^{\mathrm{S}} \in \textbf{F}^{\mathrm{S}}$ is defined by the oriented seam segment category $s_{i,j}$ and the parameters $x, y, \hat{w}, \hat{h}$ of the bounding box, i.e. $\textbf{\textit{f}}^{\mathrm{S}}=[s_{i,j}, x, y, \hat{w}, \hat{h}]$, where $x, y$ denote the coordinate of the center of the bounding box. 
$\hat{w}, \hat{h}$ denote the width and height of the bounding box. 

This bounding-box-based categorization of the seam line segment is used as the labeling rule for the SLE shown in Algorithm~\ref{alg:QOL}. To implement the SLE, we introduce:
\begin{itemize}
    \item $\lambda_\mathrm{thres}$, a threshold for the width and height of the bounding box in pixels:
    Note that the computation of the loss function during network training is very sensitive to a tiny or thin object whose area is very small. 
    $\lambda_\mathrm{thres}$ is introduced in Algorithm~\ref{alg:QOL} to guarantee a minimum size of the bounding box.
    \item Data augmentation of $\textbf{\textit{f}}^{\mathrm{S}}$ with recategorization: The orientation subclass is not invariant if the image is flipped or rotated during data augmentation. 
    Recategorization is carried out using Algorithm~\ref{alg:QOL} when the image is flipped or rotated. 
\end{itemize}

\begin{algorithm}[]
    \caption{Bounding-box-based categorization of SLE}\label{alg:QOL}
    \begin{algorithmic}
        \Require $\textbf{\textit{f}}^{\mathrm{SL}}=[j, x_1, y_1, x_2, y_2], \lambda_\mathrm{thres}$
        \Ensure $\textbf{\textit{f}}^{\mathrm{S}}=[s_{i,j}, x, y, \hat{w}, \hat{h}]$
        \State $x,y,w,h=convert\_to\_boundingbox(x_1,y_1,x_2,y_2)$
        \State $\hat{w}=w, \hat{h}=h$
        \State $\textbf{if}~(w<\lambda_\mathrm{thres}~\textbf{\&}~h<\lambda_\mathrm{thres}):$
        \State $~~~~drop\_this\_label()$
        \State $\textbf{elif}~(w<\lambda_\mathrm{thres}~\textbf{\&}~h>\lambda_\mathrm{thres}):$
        \State $~~~~i=4, \hat{w}=\mathrm{int}(h/2)$ \Comment{Vertical}
        \State $\textbf{elif}~(w>\lambda_\mathrm{thres}~\textbf{\&}~h<\lambda_\mathrm{thres}):$
        \State $~~~~i=3, \hat{h}=\mathrm{int}(w/2)$ \Comment{Horizontal}
        \State $\textbf{elif}~(x_1<x_2~\textbf{\&}~y_1>y2)~\textbf{or}~(x_1>x_2~\textbf{\&}~ y_1<y2):$
        \State $~~~~i=2$ \Comment{Upward Diagonal}
        \State $\textbf{else}:$
        \State $~~~~i=1$ \Comment{Downward Diagonal}
        \State $s_{i,j}=get\_orientated\_category(i,j)$
        \State $\textbf{return}~[s_{i,j},x,y,\hat{w}, \hat{h}]$
    \end{algorithmic}
\end{algorithm}

The proposed SLE can extract both curved and straight seams as a set of seam line segments.
The predicted seam segments are superimposed onto the original observation $o_{\tau-1}$ to generate the seam-map $o^\mathrm{S}_{\tau-1}$, as shown in Fig.~\ref{fig:subclass}~(c)

\subsubsection{Seam Crossing-segment Detectors (SCDs)}
As shown in Fig.~\ref{fig:subclass}~(b), three types of seam crossing segments, namely shoulder ($c_1$), bottom hem ($c_2$), and neck point ($c_3$) of the T-shirt, are detected by using both SCD1 and SCD2 to increase the recall of the predictions. 
The inputs of SCD1 and SCD2 are the original image observation $o_{\tau-1}$ and the corresponding seam-map $o^{\mathrm{S}}_{\tau-1}$, respectively, as shown in Fig.~\ref{fig:flowchart}. 

Note that the outputs of SCD1 and SCD2 are merged considering the maximum number of each type of crossing segments according to the confidence of the predictions, since the number of each type of seam crossing segments on the T-shirt is limited.

\subsection{Decision Matrix Iteration Method (DMIM)}
\label{subsec:decisionMatrix}
The idea of DMIM is proposed based on the observations of human unfolding actions using a flinging motion:
\begin{itemize}
    \item The Combination of Seam Segment Types (CSST) of two selected grasping points affects the unfolding performance. 
    For example, grasping two points both at the shoulders or bottom hems results in a higher unfolding quality than the other CSSTs. 
    \item 
    The current step of the unfolding action affects the unfolding performance of subsequent actions, since an appropriate action simplifies the complexity of the subsequent steps. 
    To obtain a simpler configuration for subsequent actions, we intuitively select two grasping points that are empirically far apart.
\end{itemize}

Based on these observations, we extract the human skill of unfolding by scoring the performance of each CSST through human demonstrations.
The DMIM proposed in this paper represents the grasping strategy using a decision matrix. 
The performance score of each CSST is implemented by initializing the decision matrix with human demonstrations and updating it with robot executions. 
    
\subsubsection{Decision Matrix}
As shown in Fig.~\ref{fig:priority}, the proposed decision matrix $\textbf{\textit{U}}^{\mathrm{D}}$ is constructed as an upper triangular matrix. 
In this paper, six types of seam segments are considered, including shoulder, bottom hem, neck point, solid, dotted, and neckline (the inward seam line type is treated as the dotted type in this paper for simplicity). 
The $(k, l)$ element of $\textbf{\textit{U}}^{\mathrm{D}}$, $u^{\mathrm{D}}_{k,l}$, is the performance score of the combination of the $k^{th}$ and the $l^{th}$ seam segment types, and expressed as follows: 

\begin{equation}
    u^{\mathrm{D}}_{k,l} =\frac{1}{M_{k,l}}\sum_{d=1}^{M_{k,l}}R_{d,k,l}
    \label{equ:score}
\end{equation}
where $M_{k,l}$ is the total number of trials of unfolding demonstrations performed by both human and robot. 
Note that $k, l\in\{1,2,...,6\}$ and $k \geq l$.
For each trial, the average normalized coverage $R_{d,k,l}$ is expressed as follows:
\begin{equation}
    R_{d,k,l}=\frac{1}{T_{d,k,l}}\sum_{\tau=1}^{T_{d,k,l}}{\mathrm{ncov}}(o_{\tau})
    \label{equ:futureReward}
\end{equation}
where $T_{d,k,l}$ is the number of episode steps in a trial and $\mathrm{ncov}(o_{\tau})$ is the normalized coverage of the T-shirt from an overhead camera observation $o_{\tau}$ at episode step $\tau$ during a trial, similar to~\cite{flingbot, speedfolding}, and~\cite{dextairity}.
The normalized coverage is calculated by:
\begin{equation}
    \mathrm{ncov}(o_{\tau})=\frac{\mathrm{cov}(o_{\tau})}{\mathrm{cov}_\mathrm{max}}
    \label{equ:ncov}
\end{equation}
where the coverage $\mathrm{cov}(\cdot)$ is calculated by counting the pixel number of segmented mask output by the Segment Anything Model (SAM)~\cite{sam}. 
The maximum coverage $\mathrm{cov}_\mathrm{max}$ is obtained from an image of a manually flattened T-shirt. 

The decision matrix $\textbf{\textit{U}}^{\mathrm{D}}$ is used to select grasping points from the seam segments extracted by SFEM. 
Based on the score of $u^{\mathrm{D}}_{k,l}$ in the decision matrix and the types of extracted seam segments, the CSST with the highest score is selected.
If there are multiple combinations of candidates belonging to each seam segment type of the selected CSST, the most distant candidate pair is selected as the grasping points.

\subsubsection{Initialization with Human Demonstrations and Updating with Robot Executions}
\begin{figure}[tbp]
    \includegraphics[width=\linewidth]{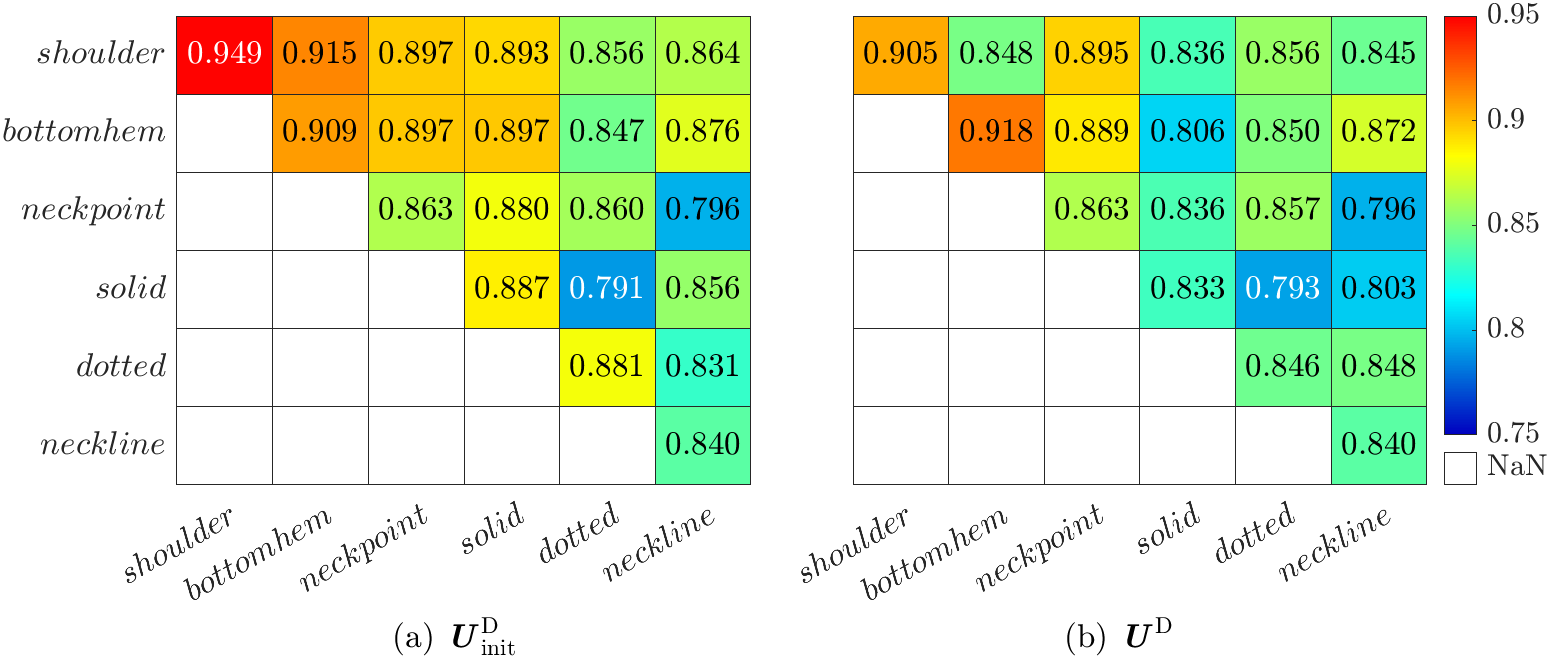}
    \caption{The decision matrices used in the experiments. 
    The matrices represent the unfolding performance score for each Combination of Seam Segment Type (CSST). 
    (a) The initial decision matrix $\textbf{\textit{U}}^{\mathrm{D}}_\mathrm{init}$ is computed with human demonstration data. 
    (b) The updated decision matrix $\textbf{\textit{U}}^{\mathrm{D}}$ is updated with data from robot trials. 
    }
    \label{fig:priority}
\end{figure}

The initial decision matrix $\textbf{\textit{U}}^{\mathrm{D}}_\mathrm{init}$, shown in Fig.~\ref{fig:priority}~(a),
is computed using only the human demonstration data by (\ref{equ:score}).
The same equation is then used to update the decision matrix with subsequent robot executions, resulting in $\textbf{\textit{U}}^{\mathrm{D}}$, as shown in Fig.~\ref{fig:priority}~(b).

Human hands exhibit more dynamic and flexible movements that allow fine-tuning of the resulting garment configuration, a capability that the robot manipulators lack. 
In other words, there is a human-to-robot gap in garment manipulation. 
To bridge this gap, we perform real robot trials to update the decision matrix.
This iterative method ensures that the grasping strategy integrates the learned human experience and the real robot explorations.

\section{Experiment}
\label{sec:experiment} 
The garment unfolding dual-arm manipulator system consists of two Denso VS087 manipulators mounted on a fixed frame, a Basler acA4096-30uc RGB camera equipped with an 8 mm Basler lens, an Azure Kinect DK depth camera, two ATI Axia80-M8 Force Torque (F$/$T) sensors, and two Taiyo EGS2-LS-4230 electric grippers. 
The robot control system is constructed based on ROS2 Humble~\cite{ros2} installed on a PC equipped with an RTX3090 GPU, an i9-10900KF CPU, and 64 GB of memory. 

\subsection{Motion Primitives Used for Experiments}
Below are the major motion primitives associated with the T-shirt unfolding experiments.

\begin{table*}[t]
\centering
\caption{Evaluation metrics of the ablation experiments}
\begin{tabular}{cccccccc}
\hline
\multirow{2}{*}{Exp. Name} & Seam Information & Matrix Iteration & \multicolumn{5}{c}{Average evaluation metric at each step over 20 trials (ncov, IoU)} \\
& (SI) & (MI) & 1 & 2 & 3 & 4 & 5 \\ \hline
SIS (Full)        &$\checkmark$& $\checkmark$& \textbf{0.707}, 0.554                  & \textbf{0.882}, \textbf{0.811}                   & \textbf{0.916}, \textbf{0.842} & \textbf{0.903}, \textbf{0.825} & \textbf{0.916}, \textbf{0.828} \\
SIS (Ab-SI)     &            & $\checkmark$& 0.590, 0.493                  & 0.765, 0.687                   & 0.850, 0.774                   & 0.863, 0.795 & 0.863, 0.756 \\
SIS (Ab-MI)     &$\checkmark$&             & 0.657, \textbf{0.574}                  & 0.764, 0.693                   & 0.823, 0.770                   & 0.822, 0.771          & 0.820, 0.764 \\
 \hline
\end{tabular}
\label{tab:ablation}
\end{table*}

\begin{figure*}[tbp]
    \centering
    \includegraphics[width=\linewidth]{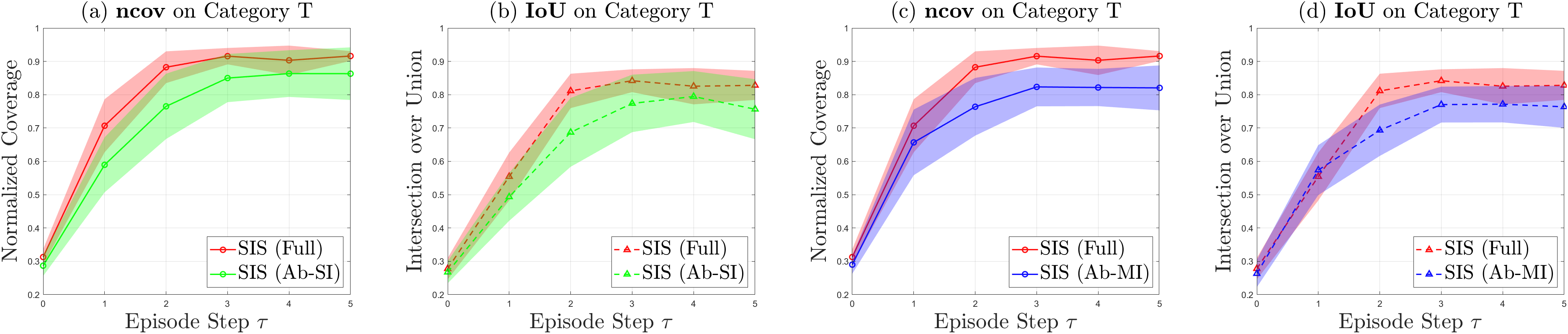}
    \caption{
    Results of ablation studies using the methods proposed in this paper. 
    The settings for each group in the plots can be found in Table~\ref{tab:ablation}. 
    We plot the normalized coverage in (a) and (c). 
    The IoU between the current configuration and the goal configuration are plotted in (b) and (d). 
    The curves in the plots denote the mean values of the metric, while the shaded areas represent the 95\% confidence intervals describing the distribution of the experimental results. 
    The data of the Full group represent the performance of our entire proposed method, while the other groups represent the performance after removing a specific module. 
    Each group undergoes a total of 20 trials of experiments using T-shirts T-1 and T-2 in Fig.~\ref{fig:clothes}, where each trial contains five episode steps.
    }
    \label{fig:ablation}
\end{figure*}

\subsubsection{Grasp\&Fling Motion Primitive}
Similar to~\cite{flingbot}, we implement a Grasp\&Fling motion primitive to speed up the process of unfolding the garment. 
After grasping, the T-shirt is stretched by the robot arms until the stretching force reaches a predefined 0.8 N using the wrist F$/$T sensors attached to both manipulators. 
The robot then generates a flinging motion that mimics the human flinging motion based on fifth-order polynomial interpolation.

\subsubsection{Randomize Motion Primitive}
To generate initial configurations of the T-shirt that are fair enough for comparison, we randomly select a grasping point from the garment area and release it from a fixed position 0.78 m above the desk using one of the robot arms. 
During the experiments described in Section~\ref{sec:ablation} and~\ref{sec:comparison}, the robot repeats this motion before each trial until the normalized coverage of the initial configuration is less than 0.4. 

\subsection{Training/Updating of SFEM and DMIM}
\subsubsection{Training of SFEM}
\label{sec:trainSFEM}
The dataset used to train the SLE consists of 328 images with manually annotated labels of seam line segments. 
Using data augmentation by image rotation and flipping, we obtain 2104 images for training and 520 images for validation with annotated labels. 
We evaluated the seam line extraction performance of the trained model using a testing set in which garments were randomly placed on the desk. We calculated the Intersection over Union (IoU) between the predicted seam line segments and the manually labeled ground truth, using a line thickness of 20 pixels. The average IoU for seam line extraction is 0.734. Considering the nature of thin objects, this IoU result is reasonable as visualized in Fig.~3~(c).

The dataset used to train SCD1 consists of 1708 original images with manually annotated labels of seam crossing segments.
The dataset used to train SCD2 consists of the same 1708 images overlaid with the seam lines extracted by the trained SLE. 
We evaluated the seam crossing detection performance of the merged outputs of trained SCDs using precision and recall. The average precision is 0.915 and the average recall is 0.878 during our experiments. 

The original images in all datasets are captured by the Basler camera. 
The resolution of the images used for training is 1280 $\times$ 1280, obtained by cropping and resizing the original images.

\subsubsection{Initialization and Updating of DMIM}
\label{sec:trainDMIM}
The decision matrix $\textbf{\textit{U}}^{\mathrm{D}}_\mathrm{init}$ is initialized with human demonstrations. 
For each human trial, we set the maximum number of episode steps to $T_{d,k,l}=5$, although two to three steps are usually enough to flatten the T-shirt. 
Once the T-shirt is flattened, we skip the remaining steps and use the same $\mathrm{ncov}(\cdot)$ as in this step for the remaining steps. 

To initialize a $u^{\mathrm{D}}_{k,l}$ for each CSST, the demonstrator is asked to randomize the T-shirt until at least one point pair in the CSST appears. 
The demonstrator then selects the furthest pair of points as grasping points.   
In the following steps of this trial, the demonstrator selects grasping points based on the demonstrator's intuition, and the $u^{\mathrm{D}}_{k,l}$ of the CSST is calculated. 
The matrix $\textbf{\textit{U}}^{\mathrm{D}}_\mathrm{init}$ is initialized as shown in Fig.~\ref{fig:priority}~(a). 
Each $u^{\mathrm{D}}_{k,l}$ is computed using (\ref{equ:score}) with the number of trials $M_{k,l}=10$.

To update the decision matrix, we conduct robot trials to bridge the human-to-robot gap in unfolding actions. 
Each robot trial consists of five episode steps. 
As shown in Fig.~\ref{fig:priority}~(b), $\textbf{\textit{U}}^{\mathrm{D}}$ is updated from $\textbf{\textit{U}}^{\mathrm{D}}_\mathrm{init}$ with (\ref{equ:score}).

\begin{figure*}[tbp]
    \centering
    \includegraphics[width=\linewidth]{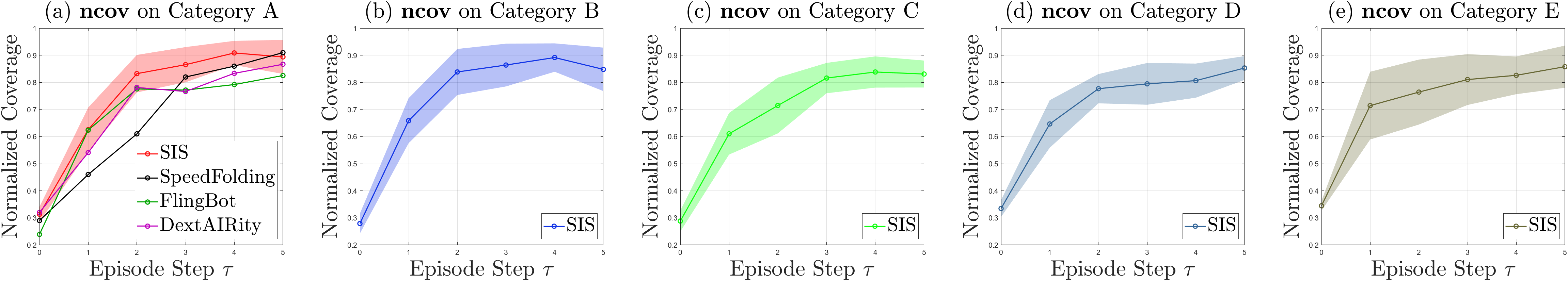}
    \caption{
    Unfolding performance evaluation on normalized coverage using Categories A, B, C, D, and E garments in Fig.~\ref{fig:clothes} and comparison with existing research using Categories A. The data we use for comparison are those presented in their original papers of FlingBot~\cite{flingbot}, DextAIRity~\cite{dextairity}, and SpeedFolding~\cite{speedfolding}. 
    }
    \label{fig:evaluation}
\end{figure*}

\subsection{Ablation Studies of the Proposed Strategy}
\label{sec:ablation}
\begin{figure*}[tbp]
    \centering
    \includegraphics[width=\linewidth]{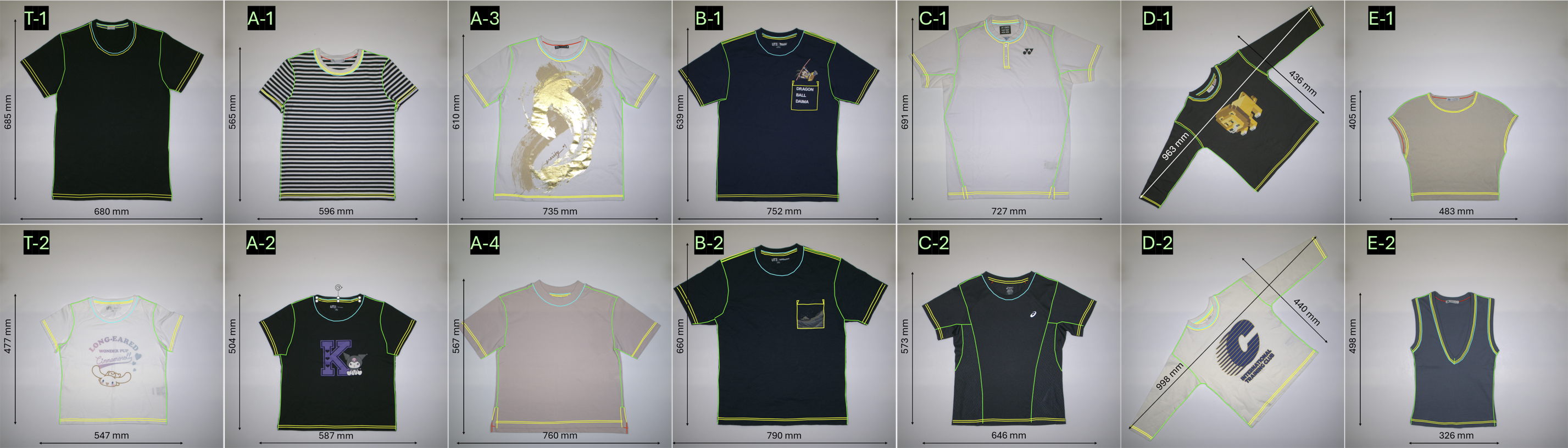}
    \caption{
    Clothes we used in the experiments. Category T is a set of 20 short-sleeved T-shirts used to train the network. T-1 and T-2 are two examples in Category T.  
    Category A contains four unseen short-sleeved T-shirts that are not part of the training dataset. 
    Categories B, C, D, and E contain unseen types of garments. 
    Category B contains short-sleeved T-shirts with a sewn pocket. 
    Category C includes two short-sleeved sports T-shirts that do not have all their seams in the contour area. 
    Category D includes long-sleeved garments.
    Category E includes tank tops.
    Seams are highlighted with colored lines. } 
    \label{fig:clothes}
\end{figure*}

In this section, we perform ablation studies for our proposed strategy.
In the following experiments, 20 trials are performed using two trained clothes in Category T in Fig.~\ref{fig:clothes}, with each trial consisting of five episode steps. 
The results are shown in Table~\ref{tab:ablation} and Fig.~\ref{fig:ablation}.
We use normalized coverage~\cite{flingbot, speedfolding, dextairity} and IoU between the current configuration and the manually unfolded configuration~\cite{clothfunnels} as evaluation metrics. 
Fig.~\ref{fig:ablation} shows the distributions of normalized coverage with 95\% confidence interval shading. 

In Table~\ref{tab:ablation}, the Full group is the complete scheme proposed in this paper based on seam information and DMIM using $\textbf{\textit{U}}^{\mathrm{D}}$. 

The Ab-SI group is carried out without the seam information extracted by the SLE and uses only the crossings detected by the SCD2 as candidate grasping points, while $\textbf{\textit{U}}^{\mathrm{D}}$ is used. 
The difference between the Ab-SI and Full groups in Fig.~\ref{fig:ablation}~(a)(b) shows that the seam information significantly improves the unfolding performance. 

The Ab-MI group is performed to show the effectiveness of the proposed iteration method. 
In this experiment, $\textbf{\textit{U}}^{\mathrm{D}}_\mathrm{init}$ is used without updating.
Fig.~\ref{fig:ablation}~(c)(d) shows the human-to-robot gap in the performance of the unfolding action.
The results of the Full and Ab-MI groups show that our proposed DMIM bridges the human-to-robot gap.

\begin{table}[tbp]
\centering
\caption{Statistical analysis results of p-values using Welch's t-test}
\begin{threeparttable}
\begin{tabular}{llllll}
\hline
\multirow{2}{*}{~Configuration} & \multicolumn{5}{c}{Episode Step}                                                   \\
 & \multicolumn{1}{c}{1} & \multicolumn{1}{c}{2} & \multicolumn{1}{c}{3} & \multicolumn{1}{c}{4} & \multicolumn{1}{c}{5}              \\ \hline
Ab-SI vs. Full &&&&& \\
~~~~~~ncov & \textbf{0.026\textsuperscript{*}} & \textbf{0.022\textsuperscript{*}} & 0.053 & 0.177 & 0.106 \\
~~~~~~IoU & 0.123 & \textbf{0.022\textsuperscript{*}} & 0.082 & 0.262 & 0.086 \\ \hline
Ab-MI vs. Full &&&&& \\
~~~~~~ncov & 0.222 & \textbf{0.013\textsuperscript{*}} & \textbf{0.004\textsuperscript{**}} & \textbf{0.016\textsuperscript{*}} & \textbf{0.007\textsuperscript{**}} \\ 
~~~~~~IoU & 0.642 & \textbf{0.009\textsuperscript{**}} & \textbf{0.018\textsuperscript{*}} & 0.089 & 0.055 \\ \hline
\end{tabular}
{\begin{tablenotes}
\item[*] ~Statistically significant performance decrease~$(p < 0.05)$
\item[**] Statistically highly significant performance decrease~$(p < 0.01)$
\end{tablenotes}}
\end{threeparttable}
\label{tab:ttest}
\end{table}

We conduct a right-tailed Welch's t-test to analyze the differences in the means of the ncov and IoU metrics before and after removing one component from the full scheme.  
Each group consists of 20 samples at each episode step. 
The null hypothesis $H_0$ posits that the Full group does not outperform the ablated configuration.
The alternative hypothesis $H_1$ posits that the Full group significantly outperforms the ablated configuration.
We reject the null hypothesis $H_0$ in favor of the alternative hypothesis $H_1$ when the p-value $p$ is less than the significance level $\alpha=0.05$, concluding that the performance of the Full group is significantly superior to that of the Ab-SI or Ab-MI group.

As shown in Table~\ref{tab:ttest}, removing the Seam Information (SI) component results in statistically significant decreases in ncov during the early steps ($p<0.05$). 
Similarly, removing the Matrix Iteration (MI) component leads to statistically significant decreases in ncov in later steps, with some decreases being highly significant ($p<0.01$). 
The IoU results exhibit similar trends.
Notably, both groups show significant performance drops during the second step, highlighting the advantages of our proposed method in addressing challenging scenarios, particularly when the garment is in a crumpled state.

Note that the following hardware-related failures are excluded from the statistical data since they are not relevant to the performance of our proposed strategy. 
\begin{itemize}
    \item Grasping failures: We consider a grasp to have failed if either one or both of the grippers failed to grasp the selected grasping points on the T-shirt. 
    We also consider the grasp to have failed if the T-shirt falls off the gripper(s) during the robot motion.
    Note that similar to~\cite{flingbot}, we have not filtered out the data when the T-shirt is grasped in its crumpled state, which makes the flinging motion ineffective. 
    \item Motion failures: These occur when the grasping points exceed the working space of the robot, or when an emergency stop is triggered due to robot singularity. 
    \item Releasing failures: We consider a garment release to have failed if the garment remains in the gripper after the gripper fingers open.
\end{itemize}

\subsection{Evaluation of unfolding performance}
\label{sec:comparison}

To evaluate the generalization ability of our proposed scheme, we conducted experiments with four unseen short-sleeved T-shirts in Category A in Fig.~\ref{fig:clothes}. 
In Fig.~\ref{fig:evaluation} (a), we plotted the average result of normalized coverage in 20 unfolding trials, each with five episode steps. 
Five continuous unfolding trials are performed for each of the four T-shirts. 
On average, the normalized coverage reaches over 0.865, 0.909, and 0.894 within three, four, and five steps, respectively. 

We compare the proposed method in this paper with those presented in~\cite{flingbot}\cite{speedfolding}\cite{dextairity}, utilizing normalized coverage as the evaluation metric.
As shown in Fig.~\ref{fig:evaluation}~(a), our SIS scheme outperforms existing methods. 
Note that our approach is simpler by using only a Grasp\&Fling motion, while SpeedFolding needs additional motions such as Pick\&Place and Drag in later steps. 
In addition, the proposed SIS aligns the T-shirt to a specific goal configuration, which SpeedFolding does not consider. 
We use IoU of resulting configuration with goal configuration to evaluate the orientation alignment performance on Category A.
Our proposed SIS reaches 0.799, 0.826, and 0.831 average IoU within three, four, and five episode steps, respectively. 

\begin{figure}[tbp]
    \includegraphics[width=\linewidth]{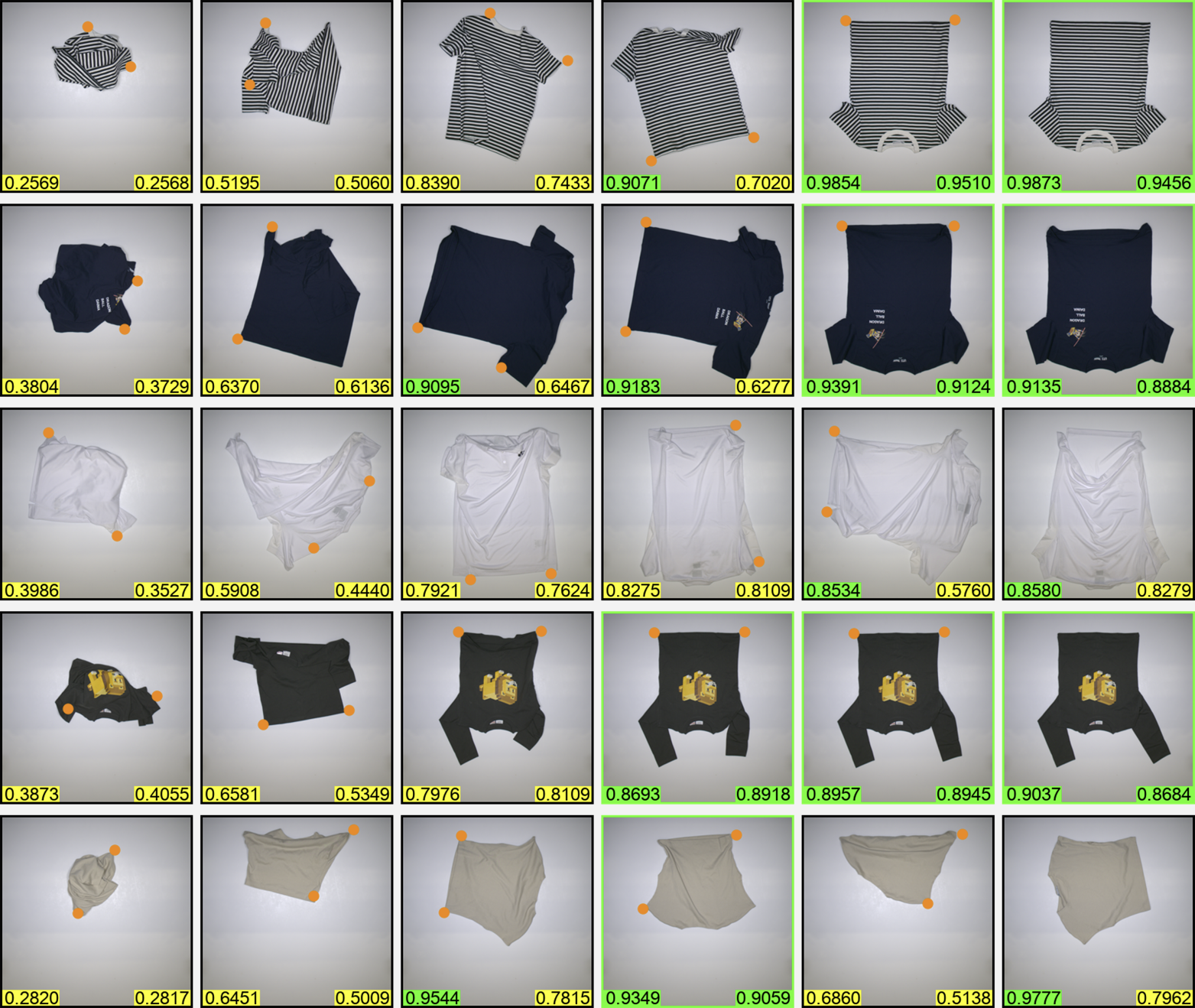}
    \caption{
    Examples of experimental results. 
    Due to space limitations, for each unseen category we show only the first of its continuous trials, each consisting of five episode steps.
    The orange dots indicate the selected grasping points for the dual-arm robot. 
    For each image observation, the number in the lower left corner shows the normalized coverage of the T-shirt, while the number in the lower right corner shows the IoU result. 
    Numbers are colored green if the metric has exceeded a threshold of 0.85. 
    If both metrics exceed the threshold, the step is marked with a green border.
    }
    \label{fig:expSteps}
\end{figure}

In Fig.~\ref{fig:evaluation}~(b) and (c), we have also plotted the normalized coverage of Category B and C. 
Category B contains two short-sleeved T-shirts with a pocket attached to the front with seams. 
Category C contains two short-sleeved sports T-shirts that have solid seams that are not in the contour area when fully unfolded. 
According to Welch's t-test, there is no significant difference in performance between experiments using Categories A and B. Therefore, we conclude that the seams on the pocket do not significantly affect the unfolding performance of our scheme.
This is because the priority of selecting the dotted seams corresponding to the pockets is low, as shown in Fig. 4. 
Even for the challenging cases of Category C, the average coverage in ten trials of our scheme reaches more than 0.83 within five episode steps, demonstrating that our scheme can also be applied to such types of sports T-shirts. 

Using the same models without additional training/updating, we conduct experiments on different types of garments, such as long-sleeved T-shirt and sleeveless shirt, shown in Category D and E in Fig.~\ref{fig:clothes}. 
A total of ten unfolding trials are performed on two cloth samples for each type of garment. 
The results are shown in Fig.~\ref{fig:evaluation}~(d) and (e). 
Although the performance suffers some degradation, our scheme can still unfold the garments without any additional retraining/updating. 
We conclude that our scheme can be applied to these two types of garments. 
Example results of unfolding trials are shown in Fig.~\ref{fig:expSteps}.

\section{Conclusion}
In this paper, we propose a Seam-Informed Strategy (SIS), which uses seam information to select a pair of grasping points for unfolding a T-shirt with a dual-arm robot system. 
Our strategy uses SFEM to extract grasping point candidates using seam information and DMIM to select the grasping points for unfolding the T-shirt while aligning its orientation in a low-cost manner. 
Experimental results using various unseen T-shirts and unseen types of garments have shown that the proposed SIS effectively handles T-shirt unfolding. 
The performance of T-shirt unfolding is promising in terms of obtaining high evaluation metrics with few episode steps.

Our future work aims to address the limitations of the current scheme when handling seamless garments, such as knitted garments, to improve the practicality and efficiency of our systems.


%





\ifCLASSOPTIONcaptionsoff
  \newpage
\fi

\vfill


\bibliographystyle{IEEEtran}
\bibliography{ref}

\vfill

\end{document}